\documentclass[11pt,a4paper]{article}

\usepackage[utf8]{inputenc}
\usepackage[T1]{fontenc}
\usepackage{amsmath,amssymb,amsfonts}
\usepackage{graphicx}
\usepackage{hyperref}
\usepackage{natbib}
\usepackage{booktabs}
\usepackage{xcolor}
\usepackage{authblk}
\usepackage{etoolbox}
\usepackage{amsmath,amssymb,amsfonts}
\usepackage{algorithm,algorithmic}
\usepackage{textcomp}
\usepackage{multirow}
\usepackage{arydshln}
\usepackage{adjustbox} 
\usepackage{pdflscape}
\begin{document}
\title{A Confounding Factors-Inhibition Adversarial Learning Framework for Multi-site fMRI Mental Disorder Identification}

\author[1]{Xin Wen\thanks{xwen@tyut.edu.cn}}
\author[1]{Shijie Guo}
\author[1]{Wenbo Ning}
\author[1]{Rui Cao}
\author[2]{Yan Niu}
\author[3]{Bin Wan}
\author[4]{Peng Wei}
\author[5]{Xiaobo Liu\thanks{xiaobo.liu@mail.mcgill.ca}}
\author[2]{Jie Xiang\thanks{xaingjie@tyut.edu.cn}}

\affil[1]{School of Software, Taiyuan University of Technology, Taiyuan, Shanxi, China}
\affil[2]{School of Computer Science (Data Science), Taiyuan University of Technology, Taiyuan, Shanxi, China}
\affil[3]{Max Planck Institute for Human Cognitive and Brain Sciences, Leipzig, Germany}
\affil[4]{Department of Psychiatry \& Behavioral Sciences, Stanford University, Stanford, CA, USA}
\affil[5]{Montreal Neurological Institute, McGill University, Montreal, QC, Canada}


\maketitle

\begin{abstract}
In open data sets of functional magnetic resonance imaging (fMRI), the heterogeneity of the data is typically attributed to a combination of factors, including differences in scanning procedures, the presence of confounding effects, and population diversities between multiple sites. These factors contribute to the diminished effectiveness of representation learning, which in turn affects the overall efficacy of subsequent classification procedures. To address these limitations, we propose a novel multi-site adversarial learning network (MSalNET) for fMRI-based mental disorder detection. Firstly, a representation learning module is introduced with a node information assembly (NIA) mechanism to better extract features from functional connectivity (FC). This mechanism aggregates edge information from both horizontal and vertical directions, effectively assembling node information. Secondly, to generalize the feature across sites, we proposed a site-level feature extraction module that can learn from individual FC data, which circumvents additional prior information. Lastly, an adversarial learning network is proposed as a means of balancing the trade-off between individual classification and site regression tasks, with the introduction of a novel loss function. The proposed method was evaluated on two multi-site fMRI datasets, i.e., Autism Brain Imaging Data Exchange (ABIDE) and ADHD-200. The results indicate that the proposed method achieves a better performance than other related algorithms with the accuracy of 75.56±1.89 \% and 68.92±5.40 \% in ABIDE and ADHD-200 datasets, respectively. Furthermore, the result of the site regression indicates that the proposed method reduces site variability from a data-driven perspective. The most discriminative brain regions revealed by NIA are consistent with statistical findings, uncovering the "black box" of deep learning to a certain extent. MSalNET offers a novel perspective on the detection of multi-site fMRI metal disorders, specifically in the context of site regression against disease detection. Moreover, it considers the interpretability of the model, which is a crucial aspect in deep learning. 
\end{abstract}

\textbf{Keywords:} 
Multi-site, Functional Connectivity, Adversarial Learning, Interpretability

\section{Introduction}
In recent decades, advances in neuroscience have significantly increased the application of non-invasive neuroimaging techniques for investigating brain functions. Among them, fMRI \cite{kliemann2024resting,jain2024age}, in particular, has become a prominent tool for classifying neurological and psychiatric disorders due to its convenience in data acquisition and minimal patient involvement\cite{hull2017resting}. Functional connectivity (FC), which assesses statistical correlations between blood oxygenation level-dependent (BOLD) signal fluctuations across spatially distributed brain regions, serves as an objective biomarker for diagnosing conditions such as Autism Spectrum Disorder (ASD) and Attention Deficit Hyperactivity Disorder (ADHD) \cite{hull2017resting,huang2020self,liang2022multi}. However, FC-based studies often rely on single-site data, limiting classification accuracy due to statistical confounding, site-specific noise, and limited generalizability. Incorporating multi-site fMRI datasets \cite{li2020multi} to form large-scale neuroimaging repositories can address these limitations, reducing problems related to small sample sizes, model overfitting, and low test-retest reliability due to acquisition noise \cite{DBLP:journals/cbm/CuiDSZZPTL23}. Thus, developing robust analytical approaches suitable for multi-site fMRI data is essential for generating reliable, reproducible, and generalizable findings, thereby advancing psychiatric research.

Several preprocessing-phase methods, such as the empirical Bayesian-based ComBat approach \cite{johnson2007adjusting}, Subsampling Maximum-mean-distance distribution alignment (SMA) \cite{mueller2005alzheimer}, and generalized linear models (GLM) \cite{nelder1972generalized}, have been proposed to mitigate site effects. However, these methods may require prior information that might not always be available, and they can perform inadequately when handling nonlinear relationships inherent in FC data. Deep learning (DL), capable of handling large-scale and complex datasets, has naturally emerged as a promising solution for multi-site fMRI analyses. Methods such as convolutional neural networks (CNN) \cite{sherkatghanad2020automated} and autoencoders (AE) \cite{eslami2019asd,zhang2023detection} have demonstrated strong capabilities in extracting discriminative features for neurological and psychiatric disorders, including ASD, Alzheimer’s disease (AD), and major depressive disorder (MDD). Despite promising results, most existing methods do not explicitly address multi-site variability.

Inspired by generative adversarial network (GAN) \cite{2014Conditional} training strategies, we propose an adversarial learning network for multi-site fMRI analysis (MSalNET) designed explicitly to mitigate site-specific effects. To our knowledge, this is the first end-to-end multi-site analytical framework that directly uses FC as input. The critical challenges of our proposed method involve accurately extracting site-level features from FC data and effectively eliminating site confounding effects during training without compromising model convergence.

Our MSalNET consists of three modules: a representation learning module, an adaptive site feature extraction module, and an adversarial learning module comprising site feature regression and primary classification tasks. These two components are trained alternately through an adversarial strategy, effectively mitigating site-specific confounding. Experimental results across multi-site datasets demonstrate that MSalNET achieves classification accuracies of 75.56±1.89\% for ABIDE-I and 68.92±5.40\% for ADHD-200. Additionally, MSalNET exhibits robust generalization capability across diverse psychiatric multi-site fMRI datasets, thus improving computer-aided diagnostic performance.

The primary contributions of this study include:
\begin{itemize}
\item An adaptive site-level feature extraction method capable of extracting generalizable site-level information directly from FC without requiring prior non-imaging knowledge.

\item A representation learning framework leveraging node information aggregation (NIA) filters to identify intrinsic functional patterns from FC data. Additionally, the designed architecture utilizes instance normalization instead of traditional pooling layers, facilitating interpretable identification of critical brain regions via backward mapping to FC.

\item An adversarial training strategy that balances site feature regression against disease classification, enhancing the generalizability and robustness of the model across diverse multi-site datasets.
\end{itemize}

The proposed MSalNET represents an effective harmonization approach applicable across various neuroimaging studies, promising new insights for future neuroscience research. The remainder of this paper is structured as follows: Section 2 details the materials and methodology, Section 3 presents extensive experimental validations, Section 4 discusses the impact of key methodological components and identifies associated regions of interest (ROIs), and Section 5 concludes the paper.

\begin{landscape}
\begin{table*}[ht]
\caption{Demographic and device information of all sites in ABIDE-I}
\label{table}
\centering
\begin{tabular}{c c c c c c c c c}
\toprule
\multirow{2}{*}{Site} & \multicolumn{4}{c}{Demographic Information} & \multicolumn{4}{c}{Acquiring Information} \\ 
&Number of Subjects & Age & Male/Female & Patient/Control & Voxel Size (mm$^3$) & Flip angle (deg) & TR (ms) & TE (ms) \\ \midrule
NYU	    &184	&6.5-39.1	&147/37	 &79/105&3.0*3.0*4.0	 &90	&2000	 &15\\
UM	    &145	&8.2-28.8	&117/28	 &68/77	&3.438*3.438*3.0  &90	&2000	&30  \\
UCLA    &109	&8.4-17.9	&96/13	&62/47	&3.0*3.0*4.0	&90	      &3000	&28\\
USM	    &101	&8.8-50.2	&101/0	&58/43	&3.4*3.4*3.0	&90	      &2000	&28\\
Leuven	&64	    &12.1-32	&56/8	&29/35	&N/A	        &90	      &1667	&33\\
Pitt	&57	    &9.3-35.2	&49/8	&30/27	&3.1*3.1*4.0	&70	       &1500	&25\\
MaxMun	&57	    &7-58	   &50/7	&24/33	&3.0*3.0*4.0	&80	      &3000	&30\\
Yale	&56	    &7-17.8	   &40/16	&28/28	&3.4*3.4*4.0	&60	     &2000	&25\\
KKI	    &55	    &8-12.8	    &42/13	&22/33	&N/A	         &75	&2500	&30\\
Trinity	&49	    &12-25.9	&49/0	&25/25	&N/A	         &90	&2000	&28\\
Standford &40	&7.5-12.9	&32/8	&20/20	&3.125*3.125*4.5	&80	  &2000	&30\\
Caltech	&38	    &17-56.2	&30/8	&19/19	&3.5*3.5*3.5	&75	     &2000	&30\\
Olin	&36	    &10-24	    &31/5	&20/16	&3.4*3.4*4.0	&60	  &1500	      &27\\
SDSU	&36	    &8.7-17.2	&29/7	&14/22	&N/A	       &90	    &2000	&30\\
SBL	    &30	    &20-64	    &30/0	&15/15	&N/A	       &80	    &2200	&30\\
OHSU	&28	    &8-15.2	    &28/0	&13/15	&3.8*3.8*3.8	&90	     &2500	&30\\
CMU	    &27	    &19-40	    &21/6	&14/13	&3.0*3.0*3.0	&73	   &2000	&30\\ \bottomrule
\end{tabular}

\end{table*}
\end{landscape}
\section{Method and Materials}
\subsection{Datasets}
We conduct a series of experiments on the ABIDE-I\cite{di2014autism}, ADHD-200\cite{adhd2012adhd} public datasets to validate the effectiveness of MSalNET. ABIDE-I dataset contains rs-fMRI data and scale information from 17 sites, with a total of 1112 subjects, including 539 ASD patients and 573 normal controls (NC). We screened 1035 subjects from official website which consist of 505 ASD subjects and 530 NC subjects with complete scale information, and the male to female ratio is 878: 157. ADHD-200 dataset consists of 8 sites with publicly available neuroimaging data from children and adolescents with 362 ADHD subjects and 585 NC. The 947 subjects consisted of structural and resting-state functional MRI data as well as scale information. Following the official guidance, we selected fMRI data from 5 sites, namely New York University Child Study Center (NYU), Peking University (Peking), Oregon Health and Science University (OHSU), Kennedy Krieger Institute (KKI), and NeuroImage (NI). Table 1 summarizes the key demographic and device details of ABIDE-I dataset. (The demographic details of ADHD-200 dataset are summarized in Supplementary Material Table S1).

\subsection{fMRI preprocessing}
The preprocessed connectome project (PCP)\cite{craddock2013neuro} was used to preprocess the ABIDE dataset. The preprocessing pipeline applies a configurable pipeline for analysis of connectomes (CPAC). Athena pipeline\cite{bellec2017neuro} is used for preprocessing ADHD-200 dataset. The preprocessing work was completed by Cameron Craddock at the Athena computer cluster of Virginia Institute of Technology. Specific preprocessing processes are seen in Supplementary Material Table S2. Both preprocessed data are available from their official website respectively.
\subsection{Functional connectivity}
A preprocessed fMRI data is a 4D time series including a 3D spatial dimension and a 1D temporal dimension. The time series were evaluated using the mean time series signal or BOLD signal of the voxels within the region of interest (ROI) in the brain atlas. The brain atlas template chosen for the ABIDE and ADHD-200 datasets was Craddock 200 (CC200)\cite{craddock2012whole} with 200 ROIs defined. The FC between the mean time series of each ROI pair was evaluated using Pearson's correlation coefficient as shown in Equation (1).

\begin{equation}
FC = \frac{\sum_{i=1}^{n} (X_i - \bar{X})(Y_i - \bar{Y})}
{\sqrt{\sum_{i=1}^{n} (X_i - \bar{X})^2} \sqrt{\sum_{i=1}^{n} (Y_i - \bar{Y})^2}}
\end{equation}
Where n is the length of time series,  and  are two time series,  and  are the mean of time series  and , respectively.
\subsection{Method}
The MSalNET is designed with two tasks (Figure 1). The first one is site feature regression, which is used to train a regression model to reflect the site variance information contained in the features learned from the representation learning module; the second one is adversarial learning, whose objective function is a combination of classification loss and site feature regression loss with an adversarial relationship. The site feature learning module is trained separately from the other two modules because the site feature vectors need to be obtained as labels for the two tasks to perform the model training under adversarial learning module. The MSalNET is trained by alternating iterations of the two tasks, and the parameters updated for each task are independent of each other without overlap, allowing the method to effectively remove the effect of site confounding on the feature extraction of the representation learning module, achieve the adversarial multi-site classification, and successfully reach convergence on the target task and Nash equilibrium. The model input is FC constructed by fMRI after preprocessing.

\begin{figure*}[ht]
\begin{center}
\centerline{\includegraphics[width=\textwidth]{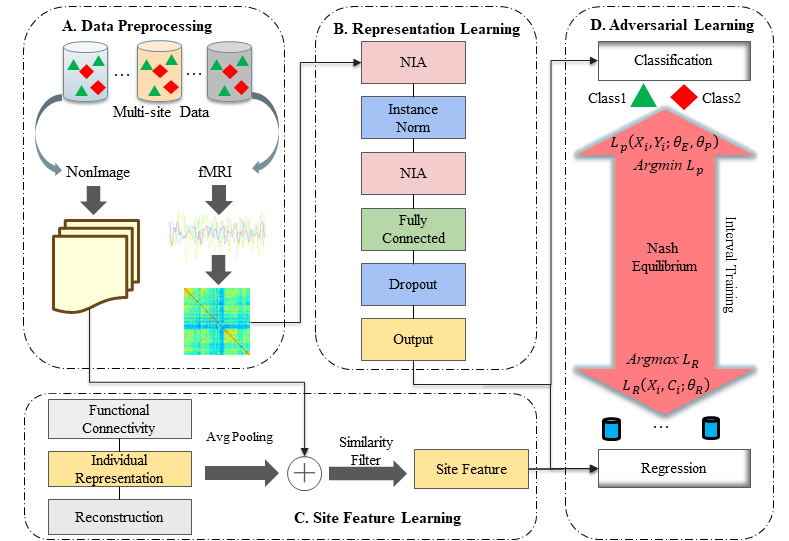}}
\caption{Overview of MSalNET. A) Data processing. For each site, FC is calculated via Pearson correlation and site level non-image information such as TR, age range and voxel size are collected for feature selection in site feature learning module. B) Representation learning module. For each FC, we design a representation learning pipeline to grab the individual level features, in the pipeline, NIA aggregates edges information to node through two 1D-CNN kernels sequentially from horizontal and vertical directions. C) Site feature learning module. Based on FC, we construct autoencoder to extract hidden representations of individuals and site features are described by average pooling forming this representation from different site ranges. In site feature learning module, similarity filter step is optional because it may not be executed due to the lack of non-image information. D). Adversarial Learning. A GAN-like adversarial learning network is designed to balance the tradeoff between site regression and disease classification. These two tasks are trained alternately, we designed a new loss function to reduce the site regression and enhance classification at the same time. }
\label{fig1}
\end{center}
\end{figure*}

\subsubsection{Representation learning}
The input of the representation learning module is FC, which is significantly different from the images in computer vision field, mainly in that the former is the non-Euclidean space while the latter is the Euclidean space, so the feature extraction effect of the conventional CNN design on FC may be unsatisfactory. Therefore, considering the topology of FC, NIA is designed as representation learning module with one-dimensional convolutional kernel performing convolutional operations on FC. NIA consists of two convolutional layers and one fully connected layer (Figure 2).

The first layer of the representation learning module uses a horizontal convolutional kernel of the form \(64@1*200\), where \(1*200\) represents the kernel size, 200 is the number of ROI and 64 is the number of channels. Since each row of the FC indicates the correlation between a certain region and the remaining regions, the \(1*200\) kernel ensures that each convolutional operation is performed on the target region with the rest regions, so a convolutional result can be seen as a feature of a certain region. After the first convolutional operation, we get 64 feature diagrams extracted from 200 ROIs of the form \(64*200*1\). The second convolutional layer uses vertical kernels to extract whole-brain features. The kernel is of the form \(128@200*1\) which convolves the features of 200 ROIs to get the whole brain features with a dimension of 128. Finally, the features are further extracted by a fully connected layer to better perform the classification task. The convolution calculation is shown in Equation (2), where \(f\) is the activation function, $x_j^m$ is the output of the current layer, $x_j^{m-1}$ is the output of the previous layer, $k_{ij}^l$ is the parameter of the current layer, \(\times\) is the dot product operation of the corresponding receptive field, and $b_{j}^m$ is the bias of the current layer.
\begin{equation}
x_j^m = f({\sum\nolimits_{i \in M_j}x_i^{m-1} \times k_{ij}^m+b_j^m})
\end{equation}

Considering that the value range of the FC is -1 to 1, so the representation learning module selects the activation function Tanh with the same value range. The Softmax function is used in the classification layer to calculate the probability of each category. The dropout layer and L2 regularization are added to prevent overfitting, and the InstanceNorm layer is applied to normalize the features in each channel.

To verify the effectiveness of NIA, we set up two feature extractors 2D-CNN and MLP to compare the classification results with NIA. The inputs of NIA and 2D-CNN are two-dimensional FC, while the inputs of MLP are one-dimensional vectors with length of 19900, obtained by removing the main diagonal from the upper triangular part of the two-dimensional FC and then flattening it.
\begin{algorithm}
\caption{Adaptive site feature learning}\label{alg:alg1}
\begin{algorithmic}
\STATE 
\STATE \vspace{0.2cm}{\textsc{Input: } Training set \(X\),\(X\) is the FC of all subjects}  
\STATE \vspace{0.2cm}{\textsc{Output: } Site feature vectors} 

\STATE \vspace{0.2cm}$ \textbf{1:}$ \hspace{0.35cm}$ \textbf{Initialization: }${loss, weights and biases} 
\STATE \vspace{0.2cm}$ \textbf{2:}$ \hspace{0.35cm}$ \textbf{While($loss$ $>$ $\varepsilon$)}$
\STATE \vspace{0.2cm}$ \textbf{3:}$ \hspace{0.7cm} Forward propagation:
\STATE \vspace{0.2cm}\hspace{2.7cm}\(X^{'}=\tanh(\text{ReLU}(X \times W + b))\)

\STATE \vspace{0.2cm}$ \textbf{4:}$ \hspace{0.7cm} Calculate the loss: \(loss=\frac{1}{n}\sum_{i=1}^{i=n}\sqrt{(X^i - X^{'i})^2}\)
\STATE \vspace{0.2cm}$ \textbf{5:}$ \hspace{0.7cm} back-propagation: \(argmin_{W,b}loss+ \lambda\parallel{W}\parallel^2\)
\STATE \vspace{0.2cm}$ \textbf{6:}$ \hspace{0.35cm}$ \textbf{end while }$
\STATE \vspace{0.2cm}$ \textbf{7:}$ \hspace{0.35cm}\textbf{Site average pooling: }\\
\STATE \vspace{0.3cm}\hspace{1.7cm}\(Z = \frac{1}{n} \sum_{i=1}^{n} avgpooling\_site(encoder(X_i))\)
\STATE \vspace{0.2cm}$ \textbf{8:}$ \hspace{0.35cm}\textbf{Feature selection: }\(C=similarity(Z, X_{scales})\)
\STATE \vspace{0.2cm}$ \textbf{9:}$ \hspace{0.35cm}\textbf{Save \(Z\) and \(C\)}
\end{algorithmic}
\label{alg1}
\end{algorithm}
\subsubsection{Multi-site feature extraction}
AE is used to extract site features as shown in Figure 3. Structurally, autoencoder is a kind of feedforward neural network, which is composed of two parts: encoder and decoder. Based on this, nonlinear dimensionality reduction is completed. The encoder encodes the input data into a low-dimensional representation, as shown in Equation(3),
\begin{equation}
h_{enc} = \phi_{enc}(x) =f(W_{enc}x+b_{enc})
\end{equation}
where \(f\) is the activation function, \(W_{enc}\) and \(b_{enc}\) are the weight and bias of the encoder, respectively. The decoder reconstructs the output of the encoder back to the original input data, as shown in formula (4).
\begin{equation}
\acute{x} = \phi_{dec}(h_{enc}) = W_{dec}h_{enc}+b_{dec}
\end{equation}

AE is trained by minimizing the reconstruction error, and the loss function is the Mean Squared Error (MSE) between the input and the reconstruction result.

The above method mainly considers extracting site features through FC in each site. This purely data-driven approach may cause feature redundancy. Therefore, when the scale information of the dataset meets the condition as prior information, feature selection is also performed on the site features. The specific feature extraction method is shown in Algorithm 1. We select the top 30\% similarity features as new site features. Because more than one kind of scale information is used, a voting mechanism is adopted for feature selection. The selected scale information includes gender, age, full-scale IQ, verbal IQ, and operational IQ (the scale information of ADHD dataset is not sufficient, so feature selection experiment was not performed for the ADHD classification experiment).
\begin{figure*}
\centerline{\includegraphics[width=\textwidth]{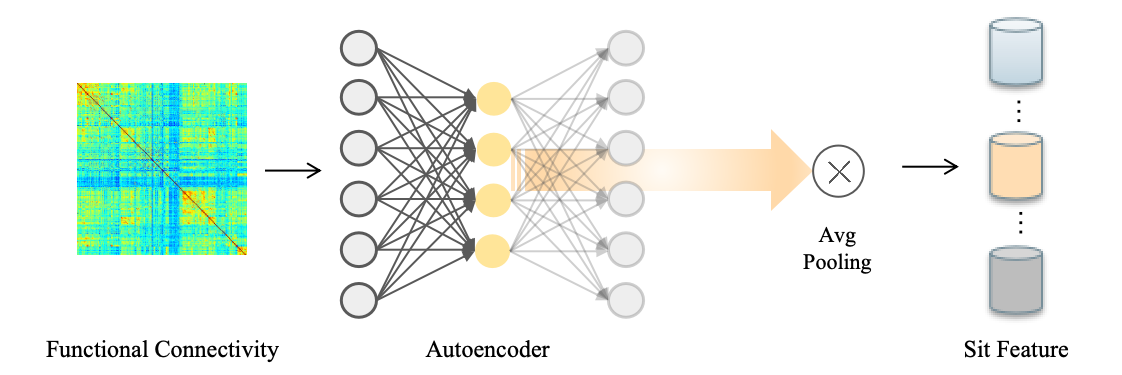}}
\caption{Structure diagram of convolutional neural network. The first convolution layer adopts a horizontal kernel and the result of each convolution can be seen as feature of a ROI. The second convolution layer adopts a vertical kernel to extract whole brain features from each ROI.}
\label{fig2}
\end{figure*}
\subsubsection{Adversarial learning}
\begin{figure*}
\centerline{\includegraphics[width=\textwidth]{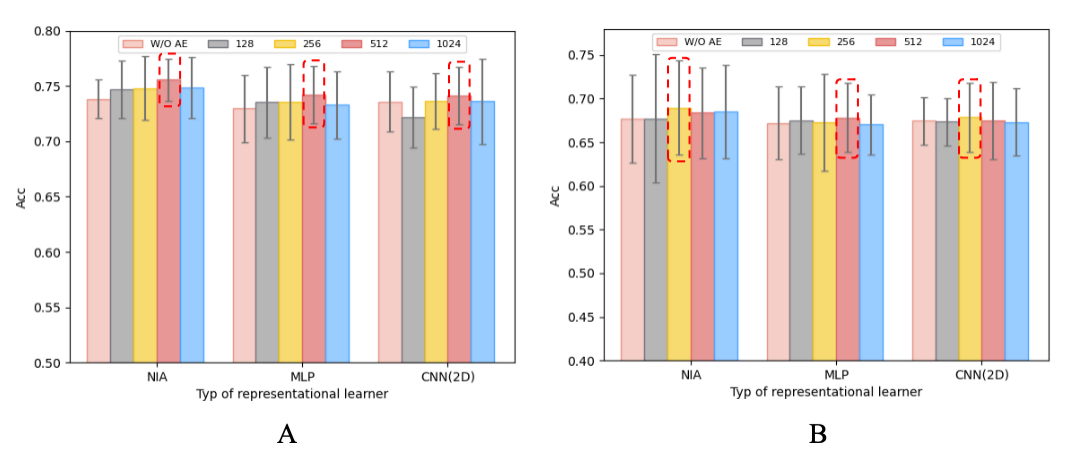}}
\caption{Site feature learning. By utilizing the characteristics of unsupervised training and nonlinear dimensionality reduction of autoencoders, the subject level feature vectors are extracted firstly. Due to the symmetry of the functional connection matrix, the lower trigonometric values, including the main diagonal, are removed. The upper trigonometric part is flattened into a one-dimensional vector and input into AE to obtain the output vector of the encoder, which is the subject level feature vector. To further obtain feature vectors that can reflect the difference information between sites, the average pooling method is used to calculate the mean vector of all subject level feature vectors in each site, which is used as the site feature vector.}
\label{fig3}
\end{figure*}
Existing methods use adversarial design to train the model in the form of a zero-sum game, that is, train the adversarial task separately to remove the confounding effect and minimize the loss. This is feasible in situations where there is only a single task such as GAN. However, when the target task and the confrontation task are inconsistent, the lowest loss value of the confrontation task may not make the target task achieve the optimal result. Therefore, we combine the adversarial task and the target task into an objective function, and train both through the loss optimization of the objective function, so that the two tasks can reach a balance in the training and ensure the optimal performance of the target task. The objective function is composed of classification and site feature regression loss functions with the purpose of reducing the classification loss and increasing the site feature regression loss. Therefore, the site feature regression in the objective function is called site feature regression confrontation. First, we introduce the symbols used by the following functions. \(X=\{X_1,...,X_N\}\), \(Y=\{Y_1,...,Y_N\}\), \(C=\{C_1,...,C_N\}\) represent the FC, labels, site labels of \(N\) subjects respectively. \(\theta_E\), \(\theta_C\), \(\theta_R\) represent the parameters of feature extractor, classifier and site feature regression component, respectively. \(P_i\) represents the probability of a category. The loss function of the site feature regression task is the MSE, as shown in Equation (5).
\begin{equation}
L_R(X_i,C_i;\theta_R) = \sum\nolimits_{i=1}^{N}{\frac{1}{m}}\sum\nolimits_{j=1}^{m}{(C_{j,i}-\acute{C}_{j,i})^2}
\end{equation}

The loss function of the main task classification is the CrossEntropy Loss, as show in Equation (6).
\begin{equation}
L_C(X_i,Y_i;\theta_E,\theta_C) = -\sum\nolimits_{i=1}^{N}[Y_i\log(P_i)+(1-Y_i)\log(1-P_i)]
\end{equation}

The model is trained in alternating iterations, with each epoch optimizing the regression task loss \(L_R\) firstly to train the regression model and then optimizing the objective function loss \(L_t\) to train the representation learning module and classifier. To achieve adversity between the two training steps, the objective function needs to minimize the classification loss \(L_P\) and maximize the regression task loss \(L_R\) at the same time. Therefore, we design the objective loss function \(L_t\) as shown in Equation (7), where \(\alpha\) is a hyperparameter, to balance the loss optimization of the two tasks and prevent \(L_R\) increases excessively leading to too strong a restriction on the representation learning module to achieve the optimal classification performance.
\begin{equation}
L_t(X_i,Y_i,C_i;\theta_E,\theta_c) = L_c(X_i,Y_i;\theta_E,\theta_c) + {\frac{\alpha}{L_R(X_i,C_i)}}
\end{equation}

The parameters of the two training updates are independent of each other. The first regression task trains the regression model to identify the site difference contained in the features extracted by the representation learning module; the second objective function is trained by optimizing both classification and regression losses in opposite directions, so that the representation learning module extracts classification features while including as few features as possible that reflect site discrepancy. Thus, during the two iterations, the regression model gradually increases its ability to identify site information, while the representation learning module extracts less site information. In this way, when the features extracted by the representation learning module are used for classification, the confounding effect of site on the classification will be weakened, and the adversarial classification is finally achieved.
\subsection{Representation interpretation}
We use back-propagation of model parameter weights to calculate the relative importance of individual ROI to find potential biomarkers for the classification of brain disorders. When designing the NIA feature extractor, we considered the one-to-one mapping relationship between the convolutional kernel and the brain region and selected a one-dimensional convolutional kernel with the same size as the number of ROIs, where each parameter of the kernel can represent the importance of a ROI. At the same time, to prevent the pooling layer from affecting the mapping relationship of back-propagation of parameter weights, the convolution process is carried out in an overlap-free receptive field, and the pooling layer is removed on the premise of ensuring low redundant features.

In the specific calculation process, the parameters of the classification layer are of the form \((N_{pre},2)\), where \(N_{pre}\) is the dimension of the classification layer and 2 is the number of categories. Here, a weight vector of the form \((N_{pre},1)\) is obtained by calculating the average of the weights of the two categories for back-propagation. The reason is that the brain regions that are important for the identification of patients and NC in the model are represented as differential brain regions, i.e., biomarkers, so that the subsequent calculation can be simplified, and the final results can be presented easily by calculating the average value. For the hidden layer, because of the MLP structure, all neurons are fully connected to each other, and the weights can be back-propagated directly by matrix multiplication, and the weights before propagation to the convolutional kernel are in the form of \((N_{C2},1)\). The second layer convolutional kernel parameters are \((N_{roi},1,N_{C1},N_{C2})\), and the matrix obtained after matrix multiplication of  kernel parameters and weight is in the form of \((N_{roi},N_{C1})\), where  is the number of ROI, \(N_{C1}\) is the number of feature maps, here also by calculating the mean value to get the importance of the dimensional brain areas.The importance of ROI of dimension \(N_{roi}\) is also obtained by calculating the mean value. Finally, the absolute value of the results is calculated and normalized to the range of 0 to 1 to represent the relative importance of different ROIs for brain disease classification. We demonstrated ROIs with brain regions ranging in importance from 0.5 to 1.
\subsection{Model Evaluation}
In the classification task, Accuracy, Precision, Recall, F1 score, and AUC are used as the evaluation indexes of the model to comprehensively investigate the classification performance of the model. These evaluation indexes are defined as below: 
\begin{equation}
Accuracy = \frac{TP+TN}{TP+TN+FP+FN}
\end{equation}
\begin{equation}
Precision = \frac{TP}{TP+FP}
\end{equation}
\begin{equation}
Recall = \frac{TP}{TP+FN}
\end{equation}
\begin{equation}
F1 = \frac{2Precision*Recall}{Precision+Recall}
\end{equation}
Where \(TP\), \(TN\), \(FP\), \(FN\) represent true positive, true negative, false positive, false negative respectively. AUC (Area Under Curve) is defined as the area under the ROC (Receiver Operating Characteristic) curve.
\section{Experiments and Results}
To evaluate the effectiveness of proposed method, we setup a comparison experiment and an ablation experiment. In all experiments, ten-fold cross-validation were performed. Each site selected 90\% of the data for the training set, and 10\% for the testing set. Subsequently, the data of the training set and testing set from all sites were integrated separately for the full-site ten-fold cross-validation. The classification performance of accuracy (ACC), area under the curve (AUC), precision, recall and F1 score are used as evaluation criteria. In comparison experiment, we compared the related methods which choose FC as input and evaluate their results involving as many sites as possible both for ASD and ADHD.  In ablation experiment, we compared different representation module, such as 2D-CNN, MLP in both adversarial training strategy and non-adversarial learning scenario. Moreover, the detail of parameter settings and the interpretation of patterns revealed via MSalNET are illustrated in following subsections. 

The proposed MSalNET could reach 75.56\% and 68.92\% ACC in ASD and ADHD respectively. The fMRI data in ABIDE are from 17 sites and those in ADHD-200 are from 5 sites, to our best knowledge, these performances are the best ones (Table 2). Compared with non-adversarial learning scenario, MSalNET shows nearly 3\% improvements in ASD and ADHD (Table 3), indicating that MSalNET not only reaches the best classification results but also has potential generalization ability. 

\subsubsection{Comparison with Related Works}
Table 2 compares the classification performance with previous DL methods dealing with ABIDE and ADHD-200 multi-site datasets. Related works mainly focus on CNN, GCN and set their efforts on representation learning. For ASD, data fed to these methods consists of whole 17 sites. The result indicated that MSalNET achieved the best classification performance. For ADHD, the accuracy reported by FCNet and DeepFMRI in the table came from each single site. In order to comprehensively evaluated the classification performance of the models on the ADHD-200 dataset, we calculated the average accuracy of each site reported in the above papers and MSalNET was superior to the above methods in terms of ADHD classification performance (moreover, MSalNET are tested from more sites).

\begin{table*}[ht]
\caption{The comparison with the existing methods for ASD and ADHD(\%)}
\label{table}
\centering
\resizebox{\textwidth}{!}{
\begin{tabular}{c c c c c c c}
\toprule
& Ref. & Sites & Method & ACC & SEN & SPE \\
\midrule
\multirow{6}{*}{ASD}        & \cite{parisot2018disease} & 17	&GCN	&70.4	&-	&- \\
                            & \cite{eslami2019asd} & 17	&ASD-DiagNet &70.3&68.3	&72.2\\
                            & \cite{liu2020attentional} & 17	&Attention selection based on Extra-Tree	&72.2	&68.8	&75.4 \\
                            & \cite{wang2021graph} & 17	 &cGCN	&71.6	&-	&- \\
                            & \cite{zhang2023detection} &17	&AE+F-score	&70.9	&70.7	&75.5 \\
                            & Ours &17	&MSalNET  	&\textbf{75.6}	&-	&- \\
                            
\cdashline{2-7}
\multirow{6}{*}{ADHD}       & \cite{riaz2017fcnet} &3	&FCNet 	&60.4	&-	&- \\
                            & \cite{riaz2018deep} &3	&DeepFMRI	&67.9	&59.1	&80.7\\
                            & \cite{wang2022contrastive} &3	&CGL-DGC	&67.0	&61.5	&72.1\\
                            & \cite{yang2023deep}  &4	&MDCN	&67.5	&72.0	&62.4\\
                            & Ours &5	&MSalNET	&\textbf{68.9} &-	&- \\
\bottomrule
\end{tabular}}
\end{table*}

\subsubsection{Ablation experiment}
In Section 2.4, we set up three representational learners for ablation: NIA, MLP, and 2D-CNN, and the classification results are shown in Table 3. For ASD, in adversarial case, NIA, MLP, and 2D-CNN, the accuracy rates were 75.56\%, 74.20\%, and 74.12\%, respectively, among which NIA had obvious advantages. In addition, when NIA was used and the hidden layer dimension of AE was set to 512, the accuracy reached 75.56\%, which was a significant improvement compared to the accuracy of 73.81\% without AE(\(p<0.05\)). In the comparison experiment with and without adversarial, when NIA was used, the accuracy increased from 72.77\% of the original one to 75.56\% of the adversarial one, and no matter what representational learning module was selected, the adversarial ones could achieve significant improvement in classification performance(\(p<0.05\)). For ADHD, accuracy rates of 68.92\%, 67.18\%, and 67.87\% were obtained using NIA, MLP, and 2D-CNN as representational learners in MSalNET, respectively. In addition, when NIA was used and the hidden layer dimension of AE was set to 256, the classification accuracy reaches 68.92\%, which is a substantial improvement compared to the accuracy of 67.67\% without AE. The accuracy improves from 65.7\% in the original ones to 68.92\% in the adversarial learning when NIA was used.
\begin{landscape}
\begin{table*}[ht]
\caption{Ablation results for MSalNET in multi-sites for ASD and ADHD(\%)}
\label{table}
\centering
\renewcommand{\arraystretch}{1.8} 
\begin{tabular}{c c c c c c c c c}
\toprule
Adversarial	&Kernel 	&Datasets	&ACC	&AUC	&Precision	&Recall	&F1Score	&Site ACC \\
\midrule
\multirow{6}{*}{\( \times \)} &\multirow{2}{*}{MLP} &ASD	&72.46±3.23	&75.70±3.43	&73.90±2.05	&71.88±3.48	&71.37±4.04	&37.04±6.50\\
                                                    & &ADHD	&65.25±3.48	&66.67±4.90	&66.76±4.30	&63.59±3.42	&62.16±4.53	&47.11±2.53\\
                            &\multirow{2}{*}{2D-CNN} &ASD	&71.37±3.45	&76.17±3.60	&73.47±3.56	&71.01±3.47	&70.41±3.91	&37.54±7.47\\
                                                    & &ADHD	&66.49±4.19	&67.45±5.43	&66.23±3.79	&64.99±4.05	&64.43±4.39	&41.58±8.62\\
                            &\multirow{2}{*}{NIA} &ASD	&72.77±3.08	&77.85±2.25	&73.59±3.17	&72.49±3.24	&72.32±3.28	&21.39±3.14\\
                                                    & &ADHD	&65.70±5.35	&66.23±8.53	&64.21±12.28	&62.79±7.46	&60.47±10.64	&43.66±1.65\\

\cdashline{2-9}
\multirow{6}{*}{\( \checkmark \)}&\multirow{2}{*}{MLP} &ASD	&74.20±3.08	&76.51±3.68	&75.55±2.72	&73.79±3.38	&73.52±3.70	&18.73±5.23\\
                                                    & &ADHD	&67.81±3.96	&66.70±4.51	&69.30±5.35	&67.49±4.77	&66.60±4.54	&31.29±2.34\\
                            &\multirow{2}{*}{2D-CNN} &ASD	&74.12±2.61	&77.10±3.15	&74.79±1.96	&73.97±2.97	&73.78±2.95	&18.58±1.89\\
                                                    & &ADHD	&67.87±3.97	&67.60±4.39	&67.92±4.10	&67.24±3.71	&67.06±3.75	&38.23±5.67\\
                            &\multirow{2}{*}{NIA} &ASD	&\textbf{75.56±1.89}	&\textbf{78.99±2.11}	&\textbf{76.60±2.25}	&\textbf{75.36±1.89	}&\textbf{75.20±2.00}	&\textbf{15.76±2.17}\\
                                                    & &ADHD	&\textbf{68.92±5.40	}&\textbf{67.66±7.25}	&\textbf{68.79±5.47}&\textbf{68.40±5.67}	&\textbf{68.15±5.77}	&\textbf{36.88±7.27}\\
\bottomrule
\end{tabular}
\end{table*}
\end{landscape}
\subsubsection{Parameters}
The hyperparameter settings and parameter searching results are shown below. The batch size was uniformly set to 10 and the early stop strategy was used in the training. To avoid overfitting, we set dropout layer and L2 regularization. The dropout rate was fixed to 0.5. In our experiment, L2, lr, \(\alpha\) in the objective function, the threshold value of site features selected according to cosine similarity and AE hidden layer dimension were used for hyperparameter search, and the value with the highest accuracy was selected as the parameter of the final model training. For ASD, L2 was selected as 0.0001, lr of the adversarial architecture was set to 0.0001, lr of the site feature extractor AE was set to 0.00001, \(\alpha\) was selected as 0.006 and AE hidden layer dimension was selected as 512, (Figure 4 A). The features in the top 30\% of the cosine similarity ranking were selected as site features (Supplementary Material Figure S1). For the ADHD classification experiments, the L2 was chosen to be 0.0001. In the training of the adversarial architecture, lr was set to 0.0001, lr of the site feature extractor AE was set to 0.00001, \(\alpha\) was chosen as 0.008, and the best AE hidden layer dimension was 256(Figure 4 B).  
\begin{figure*}[htbp]
\centerline{\includegraphics[width=\textwidth]{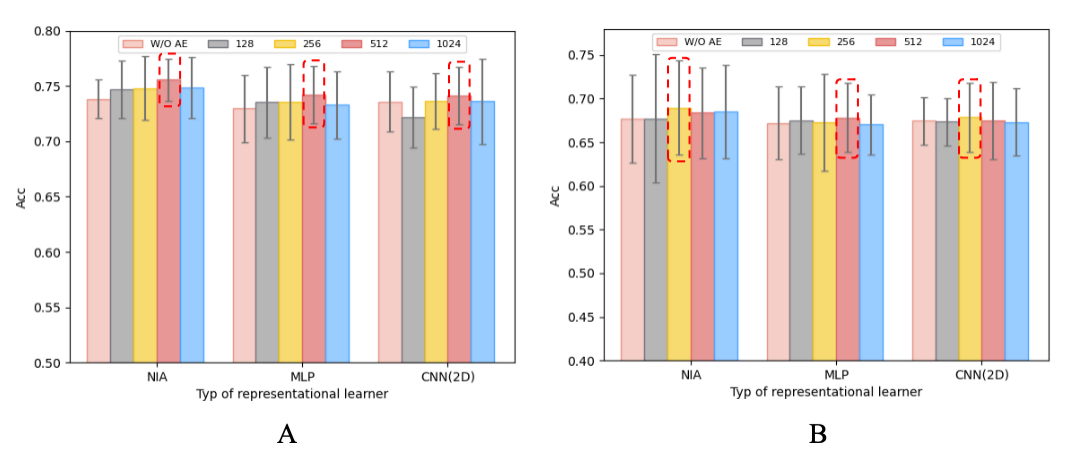}}
\caption{Classification accuracy results of site features in different AE hidden layer dimensions. W/O AE means that the site features are extracted without AE and directly obtained through FC and site average pooling, with a dimension of 19900. The numbers in the figure represent the dimensions of the hidden layer in AE and the dimension of the site feature. A represents ASD, B represents ADHD.}
\label{fig4}
\end{figure*}

When hidden layer dimension of AE was 512 and the representational learner was NIA, ASD achieved the optimal performance of model classification. Compared with the original architecture, the accuracy  increased 2.79\%. When the AE hidden layer dimension was 256, the ADHD classification accuracy reached 68.92\%, which was 3.22\% better than the original architecture. Moreover, compared without the site feature extractor, both ASD and ADHD classification accuracy were significantly improved by 1.75\% and 1.25\% respectively with AE(\(p<0.05\)), which proves that the site feature extractor could effectively extract the site key information, assist the representation learner to remove the site influence, and finally improve the classification accuracy
\subsubsection{Interpretation}
\begin{figure*}[htbp]
\centerline{\includegraphics[width=\textwidth]{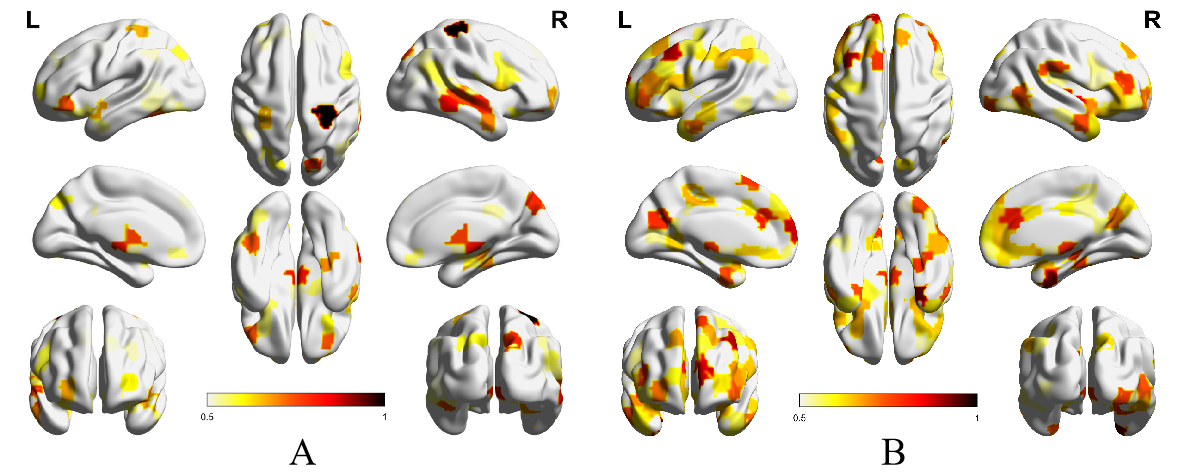}}
\caption{Classification visualization of brain region contribution with adversarial architecture. We normalize the contribution value of each brain region and select brain regions with a contribution in the range of 0.5-1 for display. The darker the color, the higher the relative importance of the brain region. A for ASD, B for ADHD.}
\label{fig5}
\end{figure*}
The contribution of important ROIs could be addressed by the method from Section 2.4(Figure 5), and a two-sample t-test between the patients and NC was performed (\(p<0.05\), FWE corrected) where the ROIs with significant FC were labeled (Figure 6)(ADHD statistical result of brain region contribution as seen in Supplementary Material Figure S2). In addition, we also compute the clustering coefficient of ROIs to compare the important ROIs revealed by the proposed method. Among them, the consistent areas were middle/inferior/superior temporal gyrus, left triangular inferior frontal gyrus, right middle frontal gyrus,  right dorsal/medial superior frontal gyrus, inferior orbitofrontal cortex, left calcarine cortex, right lingual gyrus, right fusiform gyru, right hippocampu, right parahippocampal gyrus, thalamus, right caudate, putamen, left insula, left inferior parietal lobule, right precentral gyrus, precuneus, right cuneus, left angular gyru, right superior occipital gyru, right postcentral gyrus. While the important areas only found by method explained in Section 2.5 are left middle/superior temporal pole, right middle/superior orbitofrontal cortex, right medial superior frontal gyrus, left superior medial/inferior orbitofrontal cortex, and left lingual gyru, left parahippocampal gyrus, left caudate, right insula, left cuneus.
\begin{figure*}
\centerline{\includegraphics[width=\textwidth]{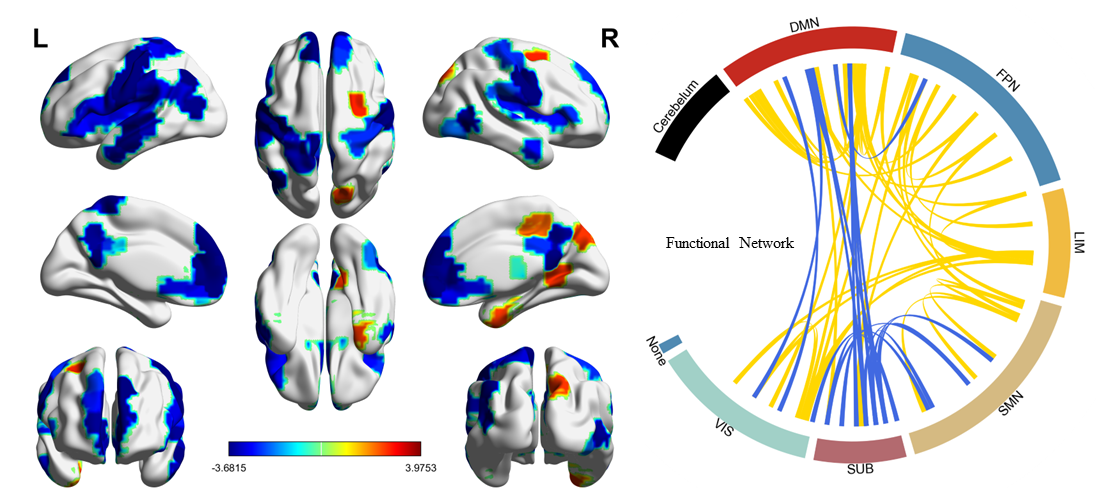}}
\caption{Left: clustering coefficient of ROIs. Red means the clustering coefficient of ASD group is higher than it in NC group. And the blue means the opposite. Right: ASD statistical result of brain region contribution. The blue lines indicate that the mean of t values in the ASD group is higher than it in NC group. The yellow lines indicate the mean of t values in the ASD group is lower than it in the NC group. }
\label{fig6}
\end{figure*}
\section{Discussion}
\subsubsection{Effectiveness of MSalNET}
According to the above results, the MSalNET has a more significant improvement in the classification performance for both ASD and ADHD diseases compared with the original one, and the classification accuracy reached 75.56\% and 68.92\%, respectively, which is superior compared with previous deep learning methods (Table 2), indicating that the adversarial brain diseases classification framework we proposed is effective in multi-site classification research. For the embedding vector obtained by original architecture, the distribution of each subject was relatively scattered. The left and right sides was caused by the difference between ASD and NC (Figure 7A). After the training of the MSalNET, the domain distribution of each sites began to gather and embed each other (Figure 7B). (The ADHD classification visualization of NIA embedding vectors is shown in Supplementary Material Figure S3)
\begin{figure*}
\centerline{\includegraphics[width=\textwidth]{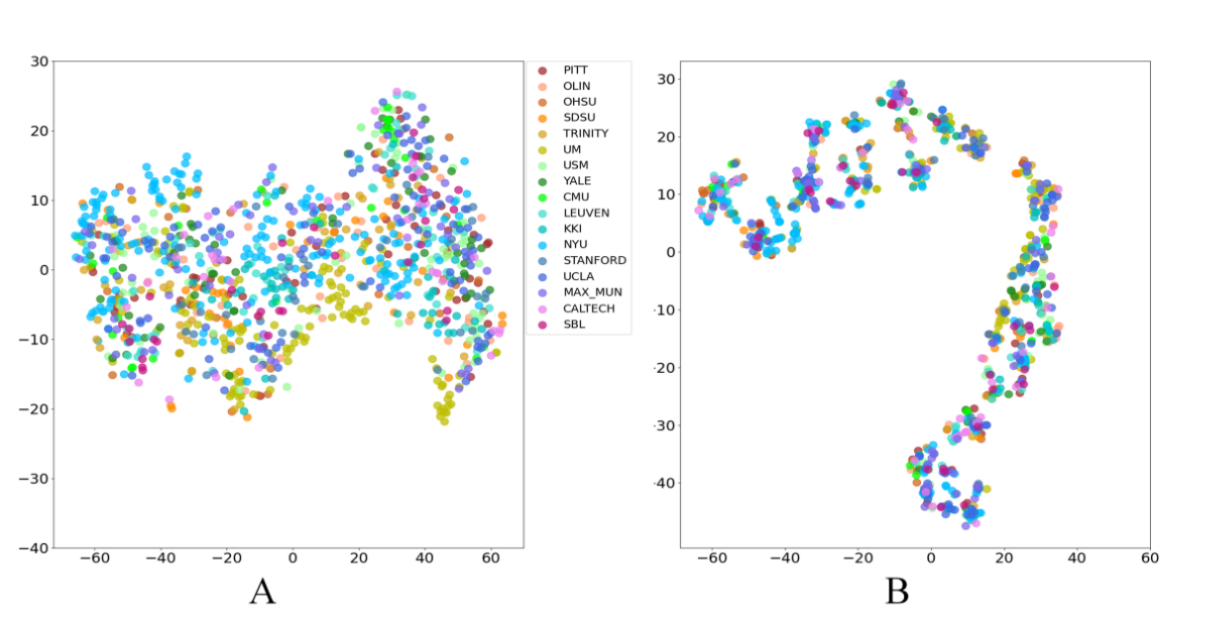}}
\caption{ASD classification visualization of NIA embedding vectors obtained using A) without adversarial, B) adversarial.}
\label{fig7}
\end{figure*}

The model trained by original architecture can achieve convergence smoothly (Figure 8A). While adapting the MSalNET, the loss value of the objective function did not converge significantly and fluctuated with a certain pattern overall, but it maintained a decreasing trend with the increase of the training times (Figure 8B). The reason is that the MSalNET uses an alternating iterative update strategy for the site feature regression loss and the objective function loss. At the beginning of each batch of training, the site feature regression loss has been trained in a round at a lower loss value, while the lower site feature regression loss will make the objective function loss larger, and subsequent training will then reduce the objective function loss. Such cyclic iterative training is reflected in the loss values as a rising and then falling process, representing the nature of confrontation. The loss of site feature regression in the objective function is cyclically decreasing and then increasing, which is exactly the opposite of the objective function loss (Figure 8C). The classification loss in the objective function still maintain the same convergence performance as the original architecture, indicating that the MSalNET does not affect the convergence of the model in the ASD classification task, and a stable classification model can still be obtained while the decline of the loss values in the training set is smoother and the gap with the testing set loss is also smaller, indicating that overfitting has been alleviated to a certain extent (Figure 8D). There is a similar pattern for ADHD classification (Supplementary Material Figure S4).
\begin{figure*}
\centerline{\includegraphics[width=\textwidth]{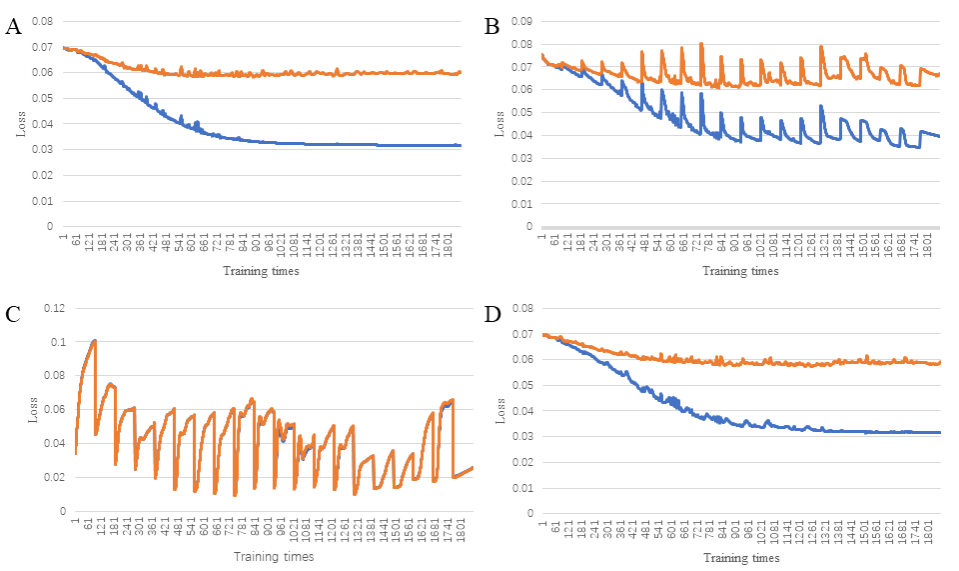}}
\caption{ASD classification model loss value.A) W/O loss. B) Adversarial objective function loss. C)Adversarial site feature regression task loss. D)Adversarial classification task loss.}
\label{fig8}
\end{figure*}

In previous studies, cross-validation and mathematical approaches were mostly used to solve the classification problem of rs-fMRI multi-site datasets to reduce the influence of site heterogeneity. In recent years, the methods widely used are generalized linear model (GLM)\cite{nelder1972generalized} and ComBat. However, GLM has limitations for nonlinear information processing\cite{reardon2021improving}. Combat coordinates fMRI data across sites in terms of statistical values, thereby reducing dataset heterogeneity and improving the ability to detect group differences, however, subsequent studies have shown limited improvement in the effectiveness of the Combat method for classifying multi-site datasets other than the original text\cite{gallo2023functional,chen2022mitigating}. While site coordination can have a large impact on the ability to distinguish domains from the data, we expect that when the dataset is balanced and the site information is independent of group membership, it may not have a large impact on the classification of interest\cite{gallo2023functional}. The method we proposed draws on the idea of GAN, which was first proposed by Ian J et al.\cite{2014Conditional} in 2014 as a new DL framework. Theoretically, any differentiable function can be used to construct the discriminative model D and the generative model G, so it can be easily combined with deep neural networks. However, D and G have difficulty achieving convergence. Thus, we designed the adversarial logic, that is, instead of using a single-task discrimination approach, we expect the site feature regression task and the classification task confront each other. The objective function controls the importance of both tasks to limit the automatic feature learning, reduce the influence of site factors on the main task classification, and ensure superior performance and convergence on the classification task. Experimental results show that the proposed network architecture can attenuate the influence of site information on representational learning and reduce site classification accuracy, while effectively improving the performance of various mental diseases classification.
\subsubsection{Backbone module}
The representation learner NIA, feature extractor AE, and adversarial architecture we adapted all have advantages in the classification of brain illness and can effectively improve the classification performance (Table 3, Figure 4). The DL models can be regarded as representational learners, of which the most commonly used is the MLP. However, MLP is too parameterized, too computationally intensive, prone to overfitting, and requires flattening 2D images, which can easily lose feature map spatial information. While 2D-CNN is more effective in the traditional computer vision domain, i.e., natural pictures\cite{li2019discover}. For our input data FC, there are obvious differences on data structure between it and natural images, mainly in that the former is non-Euclidean space while the latter are Euclidean space, and FC does not have properties such as local translation invariance that natural images have. Therefore, the feature extraction effect of traditional CNN on brain functional network may be unsatisfactory. In addition, compared with the general picture in the CV field, the global information of FC may also be important, and the FC of ROI needs to be considered comprehensively. Therefore, considering the topological structure of brain functional network, our proposed method uses NIA as the feature extraction module, and the convolution operation is performed on the brain functional network by a one-dimensional convolutional kernel. Each convolution is performed in units of brain regions and is independent of each other. There is no overlapping part of the receptive field, therefore the extracted features are less redundant. At the same time, no pooling layer is added to the NIA network, which will not affect the mapping relationship during weight back-propagation. Thus, the convolutional kernel can also improve the interpretability of the model and facilitate the exploration of the black box.

For multi-site datasets, the confounding effect is more evident in them due to the heterogeneity among sites caused by the differences in equipment, acquisition parameters, environment, and acquisition population. These confounding factors act directly or indirectly on subjects and are reflected in each MRI image, but it is more difficult to remove confounding factors directly from MRI\cite{yu2018statistical,reardon2021improving}.The representations calculated by the representational learning module in the proposed method covers functional connectivity patterns, individual differences, site information, etc. After site average pooling, according to the site grouping, the mean value of subjects features within the group is calculated to obtain site feature, which can eliminate individual differences. The site information contained in the functional connection is directly reflected on a low-dimensional basis, making it more focused as a site interference factor. For the ABIDE dataset, since its scale information satisfies the condition as a priori information, the scale information is used to calculate the cosine similarity with the feature matrix of all feature dimensions separately for feature selection, thus reducing the site feature redundancy. The ADHD dataset lacks scale information, making it impossible to perform feature selection. Nevertheless, the proposed method is sufficiently excellent for classification on the ADHD-200 dataset (Table 2).
\subsubsection{Interpretation of “black-box”}
Interpretation and explanation do not have a standard definition in current research. The former emphasizes on the degree to which the model is understood, while the latter focuses on the internal mechanism and decision logic of the model. Usually, the performance of the model with higher interpretability will be relatively poor, and it is easier to obtain intuitive measures such as feature importance through interpretable methods for structured data. Compared with regression models and decision tree, DL models do not have the ability of self-explanation, so the conventional methods in related research are hotspot map and class activation map\cite{he2016deep}. FC can be regarded as structured data to a certain extent. Our proposed method uses a NIA to perform representation learning on FC-semi-structured data, in order to take into account both model performance and interpretability and explore the “black box” together. The designed convolutional kernel size of NIA is the same as the number of ROIs, which can accomplish the one-to-one mapping relationship between convolutional kernel weights and brain regions, while no pooling layer is added to the NIA, which will not affect the mapping relationship when the weights are back-propagated. Therefore, the relative importance of each brain region to disease classification can be calculated according to the model weights through back-propagation, which is convenient for finding potential disease-related biomarkers.

By comparing the top-ranked ROIs in the contribution of ASD vs NC brain regions and the statistical results (Figure 5, Figure 6), the overlapped regions are consistent with previous studies\cite{rakic2020improving}. Among them, the insula is the source of atypical functional connectivity in ASD\cite{nomi2019insular}, and the pattern of FC between the prefrontal lobe of the DMN and the insula may be a biomarker of ASD, which is thought to be important during internal cognition and external information switching\cite{nomi2015developmental}. The strength of the temporal lobe connections to the frontal and parietal lobes is considered to be related to social skills such as language and communication, and is usually found in adults with ASD\cite{holiga2019patients}. The intensity of FC between the temporal lobe and DMN in ASD patients is lower than that of normal people, which constitutes atypical FC\cite{watanabe2017brain}. Our interpretability studies all supplement and support the conclusions of previous studies. Significant regions independently found by our interpretability method consisted mainly of left middle/superior temporal pole, dorsolateral superior frontal gyrus, orbitofrontal cortex, right insula, middle/inferior frontal gyrus, left lingual gyru, parahippocampal gyrus, cuneus, caudate. The above ROIs can be divided into three important regions, temporal pole-related regions, frontal lobe-related regions, and deep nuclei such as insula, lingual gyrus, parahippocampal gyrus, cuneus, and caudate. Among them, FC changes in the temporal pole and frontal lobe affect hearing, vision, emotion regulation and social behavior, which have been verified in ASD brain function research\cite{joshi2017integration}. Many task-based fMRI studies have shown that the functional activation of ASD patients in the insula, thalamus, cuneate lobe, and caudate nucleus has decreased\cite{blakemore2008social,kleinhans2008abnormal}. Whereas statistical methods failed to detect the FC anomalies in the regions of deep nuclei, our approach can locate them. Meanwhile, white matter-related research have shown that fiber changes on the corpus callosum are also one of the manifestations of ASD lesions, while deep nuclei such as the insula and lingual gyrus can regulate information interaction between the left and right hemispheres, fine-tune voluntary movements, and intervene in other higher cortical functions, such as motion, memory, eye movement, reward processing and motivation planning, regulation, etc\cite{ameis2012imaging,kana2012archeologist}. Integrating the statistical results with the interpretability studies showed (see Supplementary Materials for ADHD-related interpretations) that our proposed interpretability method could not only label the same important brain regions as the rs-fMRI studies, but also additionally label the brain regions reported in the task-fMRI and sMRI studies, illustrating the validity of the interpretable approach.
\subsubsection{Homogeneity in ABIDE and ADHD-200}
The proposed method was performed on the two multi-site fMRI datasets, ABIDE and ADHD-200 respectively, and the comparative experiments confirmed that the MSalNET is superior to the original ones in terms of classification evaluation indicators such as accuracy with lower site classification accuracy, indicating that the adversarial architecture we proposed weakened the influence of data heterogeneity and can achieve better performance on a variety of multi-site datasets. In the two multi-site datasets, only ABIDE provides complete scale information, thus the ADHD-200 dataset has no feature selection step, and it can be determined that the adversarial architecture and AE site feature learning in the proposed method are effective. In the contrast experiment with and without AE, we found that using AE to extract site features is more effective in improving classification accuracy than directly using FC to flatten into a one-dimensional vector with a length of 19900 as site features. It may be because AE can effectively reduce the redundant features of the original vector and retain the most distinctive characteristics of each site, so that the representation learner in the MSalNET can remove the site factor efficaciously. However, when different datasets use the AE module, the hidden layer dimensions corresponding to the optimal results are inconsistent. Therefore, we added an automatic parameter optimization mechanism in the method setting.

In the process of integrating large neuroimaging datasets, it is inevitable that subjects and site information will introduce confounding effects. For example, in some EEG epilepsy predictions, a certain number of subjects are unable to predict when subjects are independent, which is caused by individual differences\cite{kuhlmann2018seizure,gallo2023functional}. It is also conventional to remove the acquisition instrument, scan information, etc. as covariates in the fMRI preprocessing stage using GLM\cite{wang2019identifying}, and researchers have further used statistical methods such as Combat to remove site interference\cite{mueller2005alzheimer,chen2022mitigating}. Our results show a decline in site classification accuracy in adversarial structures on two multi-site datasets, confirming a certain degree and success in excluding site interference from a data-driven perspective. In addition, the MSalNET is still efficacious regardless of whether feature selection is added to the AE module, in which averaging subject features grouped by site as a site regression volume is considered effective in a data-driven perspective. Meanwhile, considering that the subjects will inevitably add common disturbances to the fMRI signal in theory under the unified acquisition equipment, parameters, alternating current and other environmental factors, averaging the representation at the site level will inevitably reflect the common representation in each site, so we can consider it as a site regressor. In this way, the artificial introduction of prior information in the direct setting of site regressions is avoided, and an end-to-end pipeline can be formed in our method, which is convenient for deployment and verification. At the same time, this may also be a worthwhile method promoting to aggregate site information from individual differences, to play a greater role in future multi-site research.
\section{Conclusion}
Aiming at the multi-site problem existing in fMRI research, we propose a new method to alternately train the disease classification task and the site classification task in two steps during the model training process with the adversarial idea to eliminate site confounding. In the proposed method, we introduced 1)node information assemble module, 2) site level feature extraction module, 3) GAN-like adversarial learning module. Moreover, we also try to reveal the “black-box” of deep learning methods in a neuroimage perspective, and the result indicate that patterns via our interpretation method could match biological explanations from another research. The proposed method reached generalization performance in different multi-site fMRI datasets which could be applied to other datasets. The comparison result indicate that the proposed method has superior performance than the existing ones. And the ablation results indicate that the proposed three mechanisms have positive effect on the whole method. Our work mainly focuses on fMRI, multi-modal data fusion is the next attempt which has been proved effective in classification tasks. In the feature level, besides FC more measures should be considered.



\bibliographystyle{plainnat}
\bibliography{main}

\end{document}